\documentclass[letterpaper]{article} %
\usepackage{aaai2026}
\usepackage{times}  %
\usepackage{helvet}  %
\usepackage{courier}  %
\usepackage[hyphens]{url}  %
\usepackage{graphicx} %
\usepackage{natbib}  %
\usepackage{caption} %
\usepackage{algorithm}
\usepackage{algpseudocode}  
\usepackage{booktabs}  
\usepackage{multirow} 
\usepackage{amsmath}  %
\usepackage{amssymb}  %
\usepackage{booktabs}   %
\usepackage{multirow}   %
\usepackage{array}      %
\usepackage[dvipsnames]{xcolor}   
\usepackage{pifont}  %
\newcommand{\cmark}{\ding{51}}  
\newcommand{\xmark}{\ding{55}}

\usepackage{etoolbox}
\makeatletter
\AfterEndEnvironment{algorithm}{\let\@algcomment\relax}
\AtEndEnvironment{algorithm}{\kern2pt\hrule\relax\vskip3pt\@algcomment}
\let\@algcomment\relax
\newcommand\algcomment[1]{\def\@algcomment{\footnotesize#1}}
\renewcommand\fs@ruled{\def\@fs@cfont{\bfseries}\let\@fs@capt\floatc@ruled
  \def\@fs@pre{\hrule height.8pt depth0pt \kern2pt}%
  \def\@fs@post{}%
  \def\@fs@mid{\kern2pt\hrule\kern2pt}%
  \let\@fs@iftopcapt\iftrue}
\makeatother

\usepackage{newfloat}
\usepackage{listings}
\DeclareCaptionStyle{ruled}{labelfont=normalfont,labelsep=colon,strut=off} %
\floatstyle{ruled}
\newfloat{listing}{tb}{lst}{}
\floatname{listing}{Listing}
\usepackage{hyperref} 
\nocopyright 

\title{\raisebox{-0.5em}{\includegraphics[height=1.8em]{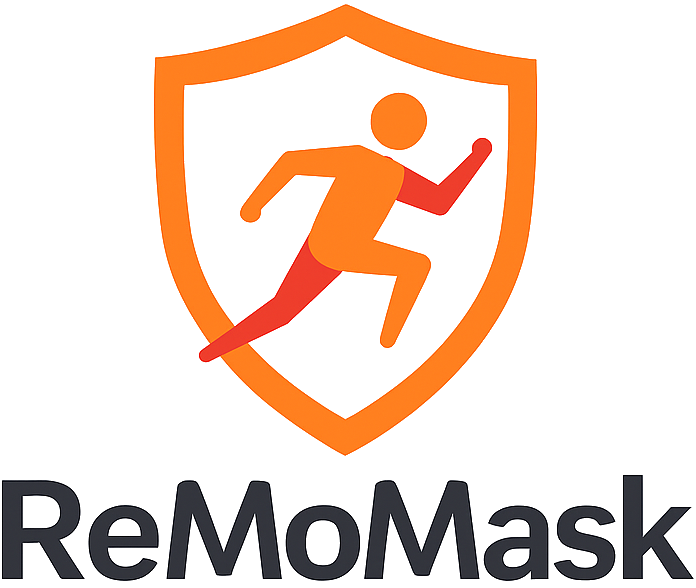}}~ReMoMask: Retrieval-Augmented Masked Motion Generation}

\author{
    Zhengdao Li$^{1*}$\quad Siheng Wang$^{2*}$\quad Zeyu Zhang$^{1*\dag}$\quad Hao Tang$^{1\ddag}$
}

\affiliations{
    $^1$Peking University\quad
    $^2$Jiangsu University \vspace{0.1cm}\\
    \scriptsize $^*$Equal contribution. $^\dag$Project lead.
    $^\ddag$Corresponding author: bjdxtanghao@gmail.com.
}

\usepackage{bibentry}

\begin{document}

\maketitle

\begin{abstract}
Text-to-Motion (T2M) generation aims to synthesize realistic and semantically aligned human motion sequences from natural language descriptions. However, current approaches face dual challenges: Generative models (e.g., diffusion models) suffer from limited diversity, error accumulation, and physical implausibility, while Retrieval-Augmented Generation (RAG) methods exhibit diffusion inertia, partial-mode collapse, and asynchronous artifacts. To address these limitations, we propose ReMoMask, a unified framework integrating three key innovations: 1) A Bidirectional Momentum Text-Motion Model decouples negative sample scale from batch size via momentum queues, substantially improving cross-modal retrieval precision; 2) A Semantic Spatiotemporal Attention mechanism enforces biomechanical constraints during part-level fusion to eliminate asynchronous artifacts; 3) RAG-Classier-Free Guidance incorporates minor unconditional generation to enhance generalization. Built upon MoMask's RVQ-VAE, ReMoMask efficiently generates temporally coherent motions in minimal steps. Extensive experiments on standard benchmarks demonstrate the state-of-the-art performance of ReMoMask, achieving a 3.88\% and 10.97\% improvement in FID scores on HumanML3D and KIT-ML, respectively, compared to the previous SOTA method RAG-T2M.
Code: \url{https://github.com/AIGeeksGroup/ReMoMask}. Website: \url{https://aigeeksgroup.github.io/ReMoMask}.
\end{abstract}

\begin{figure}[t!]
    \centering
    \includegraphics[width=\linewidth]{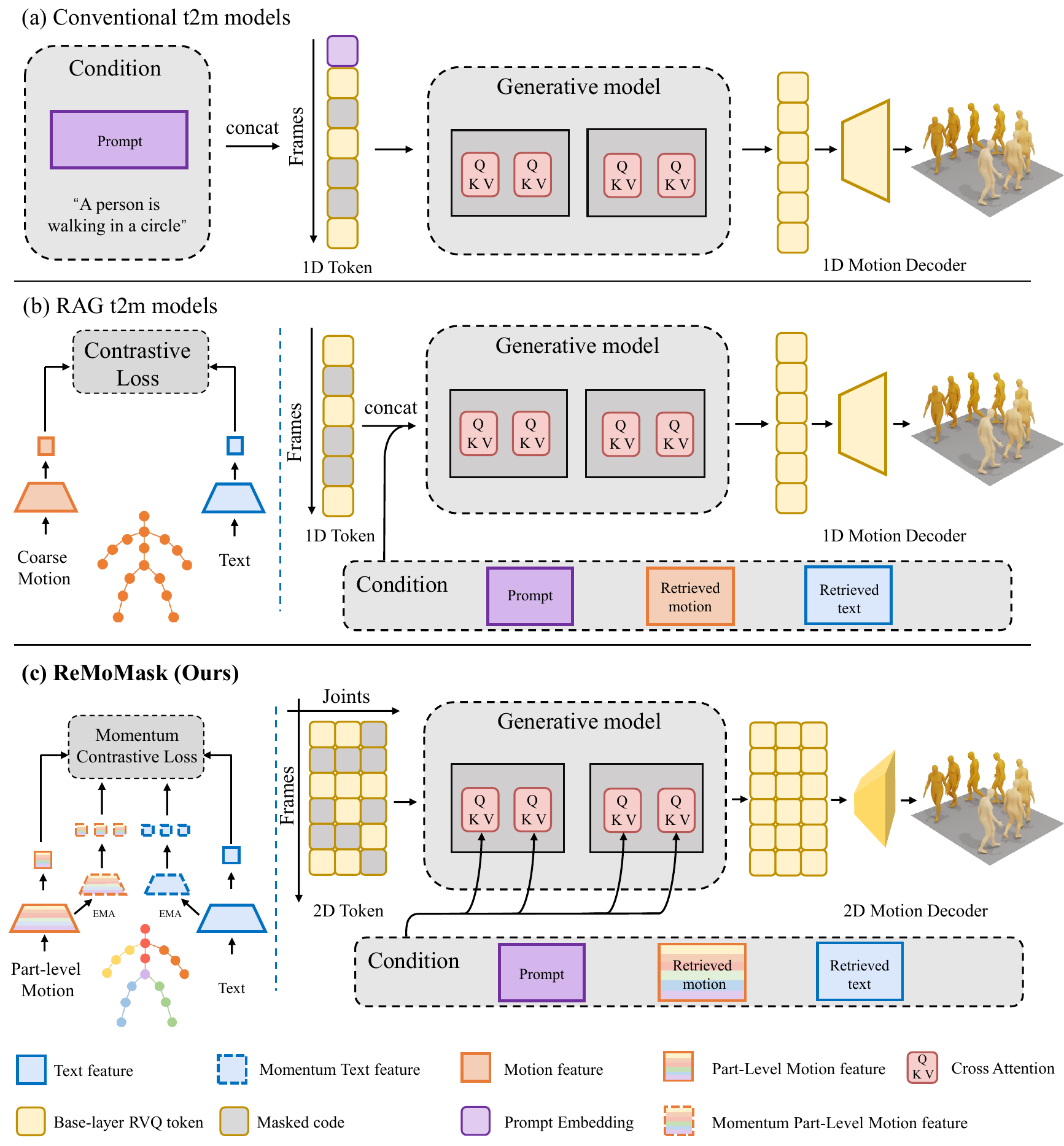}
    \vspace{-10pt}
    \caption{\textbf{Comparison between t2m models.} (a) The conventional t2m models. (b) The Existing RAG-t2m models. (c) The framework of our proposed ReMoMask.}
    \vspace{-10pt}
    \label{fig:teaser}
\end{figure}

\section{Introduction}
Human motion generation has attracted growing attention due to its broad applicability in domains such as gaming, film production \cite{zhang2024motionavatar}, virtual reality, and robotics. Recent advancements aim to synthesize diverse and realistic motions to reduce manual animation costs and enhance content creation efficiency.
Among these efforts, text-to-motion (T2M) generation \cite{zhang2024motion,zhang2024infinimotion,zhang2025motion,zhang2024kmm} has emerged as a particularly intuitive paradigm, where the objective is to generate a sequence of human joint positions based on a textual description of the motion.

Previous research on text-to-motion generation could be categorized into two directions. The first is \textit{conventional t2m models}, which have explored various generative models such as generative adversarial networks (GANS)~\cite{goodfellow2014generative}, variational autoencoders (VAEs)\cite{kingma2014auto}, diffusion models~\cite{ho2020denoising}, motion language models~\cite{zhang2023motiongpt}, or generative masked models~\cite{li2024generative}. Among them, generative mask models such as MoMask and MMM use a motion quantizer to transform a motion seuqnce into discrete tokens and train the model through randomly masked token prediction, resulting in high-fidelity motion synthesis. The second, termed \textit{RAG-t2m models}, leverages retrieved text and motion knowledge from an external database to guide the motion generation, complementing the generative model, which performs well in handling uncommon text inputs. ReMoDiffuse and ReMoGPT represent two representative approaches: the former relies on the text-to-text similarity between captions using CLIP for the retrieval, while the latter adopts a cross-modal retriever for improved motion alignment.

As shown in Figure~\ref{fig:teaser}, although conventional t2m models provide precise motion synthesis and RAG-t2m models enhance generation versatility, these approaches still face two key challenges. First, the motion retriever is trained using mini-batch, which suffers from a limited number of negative samples, thus limiting the learning of robust representations. Second, concatenating the text condition with a 1D motion token is insufficient for modeling the relationships among the text condition, motion spatio-temporal information, and retrieved knowledge. These limitations highlight the need for a new retriever training paradigm that supports larger negative samples, as well as a more powerful information fusion mechanism.

To address the first challenge, we propose a bidirectional momentum text-motion modeling (BMM) algorithm, which provides a mechanism of using two momentum encoders and maintaining dual queues that hold text and motion negative samples, respectively. At each step, the negative samples encoded by momentum encoders of the current mini-batch are enqueued, while the oldest are dequeued. This design decouples the negative pool size from the mini-batch size, allowing a larger negative set for contrastive learning. Furthermore, the two momentum encoders are updated via an exponential moving average of their online counterparts, ensuring temporal consistency across negatives. 

Moreover, to address the second challenge, we introduce a semantics spatial-temporal attention (SSTA) mechanism. Unlike previous motion VQ quantizations that produce a 1D token map and neglect spatial relationships between individual joints, SSTA tokenizes the motion into a 2D token map via a 2D RVQ-VAE, which not only captures the temporal dynamics but also aggregates the spatial information. During the later generation, the 2D token map is flattened and fused with text embedding, retrieved text features, and retrieved motion features by redefining the Q, K, V matrix in the transformer layer. Compared to simply concatenating conditions, our powerful information fusion mechanism enables comprehensive alignment across text guidance, retrieved knowledge, motion temporal dynamics, and even motion spatial structure, facilitating precision and generalization simultaneously. 

Together, these components constitute ReMoMask, an early retrieval augmented text-to-motion masked model, which outperforms prior text-to-motion approaches on HumanML-3D and KIT-ML benchmarks.

The contributions of our paper can be summarized as follows:
\begin{itemize}
    \item We propose ReMoMask, an innovative RAG-t2m masked model, equipped with a powerful information fusion mechanism, SSTA, which enables effective fusion of conditions with both temporal dynamics and spatial structure of motion. 

    \item To enlarge the negative sample pool in text-motion contrastive learning, we proposed a bidirectional momentum text-motion modeling algorithm (BMM), which decouples the number of negative samples from the mini-batch size and achieves state-of-the-art performance on text-motion retrieval.
    
    \item ReMoMask generates motion sequences with better generalization and precision than MoMask~\cite{guo2024momask}, achieving a 3.14\% improvement in MM Dist on HumanML3D and 32.35\%  in FID on KIT-ML.
\end{itemize}

\section{Related Work}

\subsection{Text-to-Motion Generation}

In the field of text-driven 3D human motion generation, numerous research achievements have been made. Initially, Text2Motion pioneered the establishment of the mapping between text and motion through adversarial learning. Subsequently, TM2T~\cite{guo2022tm2t} first introduced vector quantization (VQ), and T2M-GPT~\cite{zhang2023t2m} employed autoregressive transformers for semantic control. However, T2M-GPT suffered from error accumulation during unidirectional decoding. MoMask~\cite{guo2024momask} proposed a hierarchical residual quantization framework, decomposing the motion into base tokens and residual tokens, and combined with bidirectional masked transformers for parallel decoding, achieving remarkable results on the HumanML3D dataset. Regarding masked modeling, MotionCLIP~\cite{tevet2022motionclip} achieved unsupervised cross-modal alignment using CLIP but was limited by continuous representations.The diffusion models have significantly advanced this field. MotionGPT~\cite{motiongpt2023} discretizes motion and leverages autoregressive transformers to unify generation tasks, mitigating error accumulation. The denoising diffusion models (DDPMs) proposed by Song et al.~\cite{song2020denoising} enable high-quality parallel generation via non-autoregressive paradigms. Building on language model pretraining techniques by Radford et al.~\cite{radford2019language}, researchers further enhance text semantic control. Current methods integrate discrete representations with diffusion frameworks to balance efficiency and generation quality. Moreover, for fine-grained part-level control, ParCo introduced a breakthrough approach, it discretizes whole-body motion into six part motions (limbs, backbone, root) using lightweight VQ-VAEs to establish part priors. 

\subsection{Retrieval-Augmented Generation}

Retrieval-augmented generation (RAG) has become a powerful approach for enhancing large language models (LLMs) by incorporating external knowledge retrieved during inference~\cite{guo2025lightragsimplefastretrievalaugmented,qian2024memorag,gao2023retrieval}. Initially developed for natural language processing tasks, RAG helps models generate factually grounded, contextually relevant, and domain-specific responses, addressing common issues such as hallucinations, outdated knowledge, and limited expertise in closed models. A typical RAG system consists of three key components: indexing, retrieval, and generation. Data is first encoded and stored in a vector database; at inference, the most relevant information is retrieved based on the input query and used to guide generation.

The development of RAG builds on the evolution of information retrieval (IR) methods. Early IR systems relied on sparse vector representations, with techniques like TF-IDF~\cite{sparck1972statistical} and BM25~\cite{robertson1995okapi} ranking documents based on term frequency and inverse document frequency. These methods, however, struggled to capture semantic similarity due to their reliance on exact term matches. With the rise of deep learning, neural IR models began representing queries and documents as dense vectors, typically using a bi-encoder~\cite{karpukhin2020dense,izacard2021unsupervised} model or cross-encoder~\cite{nogueira2019passage} model. These representations enabled more effective semantic matching through similarity computations, laying the foundation for modern retrieval-based generation.

Beyond text, RAG has been extended to multimodal domains such as image~\cite{qi2025ar}, video~\cite{ren2025videorag}, and motion generation~\cite{zhang2023remodiffuseretrievalaugmentedmotiondiffusion,kalakonda2024moragmultifusionretrieval, Yu_Tanaka_Fujiwara_2025}, where external visual or motion references are retrieved to guide the generation process. These advances highlight the flexibility and broad applicability of the RAG framework across diverse tasks and modalities.

\section{The Proposed Method}
\subsection{Framework Overview}
\begin{figure*}[t] 
    \centering
    \includegraphics[width=\textwidth]{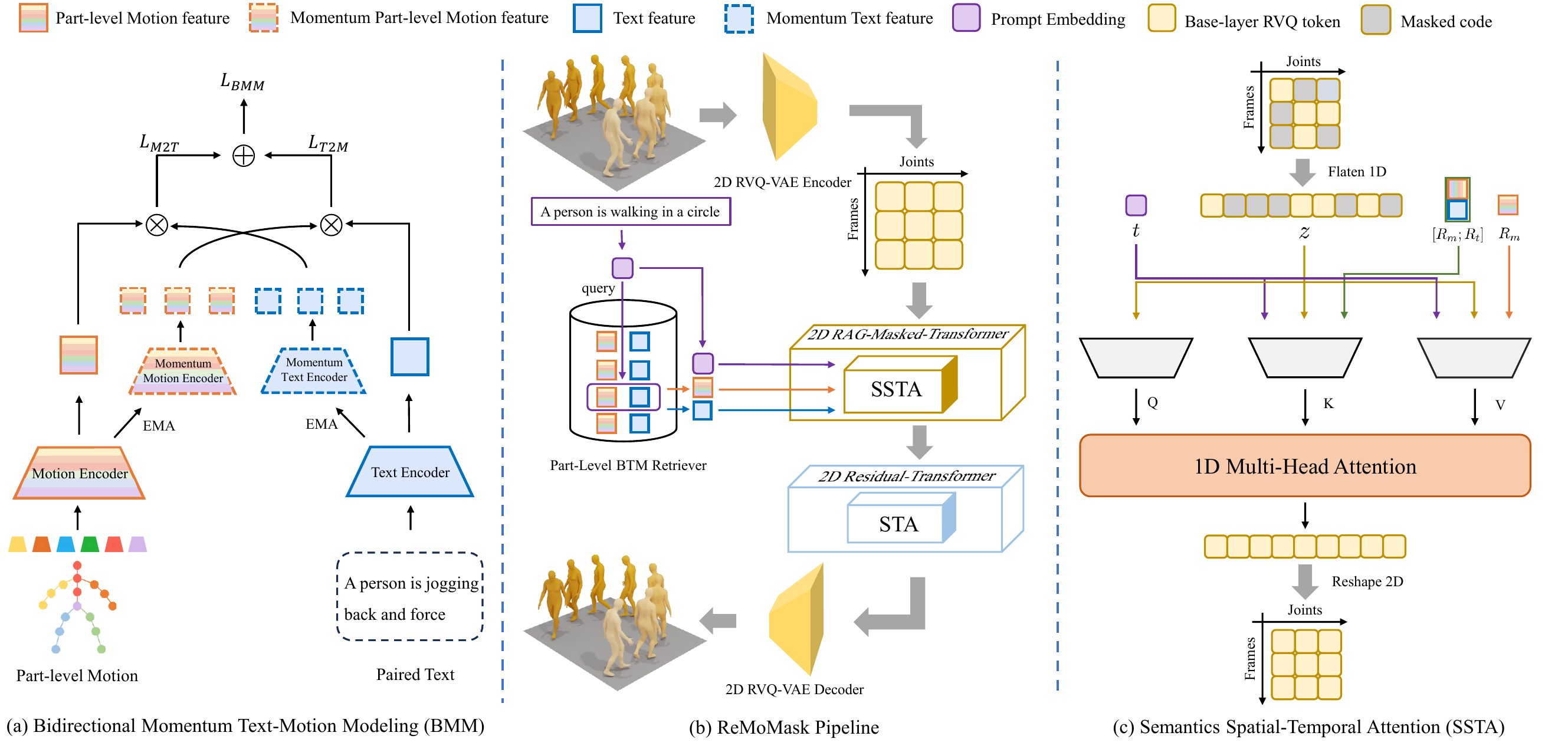} 
    \caption{\textbf{Overview of ReMoMask}. (a) Bidirectional Momentum Contrastive Retrieval (BMM) uses two momentum queues, enabling a large pool of negative samples for contrastive learning.
    (b) ReMoMask quantizes a motion sequence into a 2D token map, capturing not only temporal dynamics but also spatial structure. After that, a Part-Level BMM Retriever retrieves relevant text and motion features based on the prompt embedding. All these conditions are fused via an SSTA module in a 2D RAG-Mask-Transformer together with the latent motion representaion.
    (c) Semantic Spatial-temporal Attention (SSTA) first flattens the masked 2D token map into a 1D structure, then redefines the Q, K, V matrix utilizing the conditions above, providing effective semantic alignment between the conditions and the spatial-temporal information of motion
    } 
    \label{fig:pipeline}
\end{figure*}

Figure~\ref{fig:pipeline} shows the overall architecture of ReMoMask. 
To ensure the quality of the motion in both temporal dynamics and spatial structure, we quantize a motion sequence into a 2D spatial-temporal map via a 2D RVQ-VAE encoder. During generation, starting from an all masked 2D token map, ReMoMask first retrieves text and motion features using a \textit{Part-Level Bidirectional Momentum Text-Motion Retriever}, which is trained with the \textit{Bidirectional Momentum text-motion modeling} (BMM) algorithm to enable a large negative samples pool. These retrieved features are then fed into the Masked Transformer and fused by \textit{Semantics Spatial-Temporal Attention} (SSTA), providing strong semantic alignment and guidance for reconstructing the core motion structure. Finally, a Residual Transformer refines motion details, and the latent motion vector is decoded through a 2D RVQ-VAE decoder. 

\subsection{Bidirectional Momentum Text-Motion Modeling}

As shown in Figure~\ref{fig:pipeline}(a), we adopt a dual momentum encoder architecture equipped with corresponding memory queues. Let $f^m$ and $f^t$ denote the motion and text encoders, parameterized by $\theta^m$ and $\theta^t$, respectively. To ensure temporal consistency across negative samples, we introduce two momentum counterparts $\hat{f}^m$ and $\hat{f}^t$, with parameters $\hat{\theta}^m$ and $\hat{\theta}^t$, which are updated using exponential moving averages with momentum coefficient $\tilde{m}$:
\begin{align}
\hat{\theta}^m & = \tilde{m} \cdot \hat{\theta}^m + (1 - \tilde{m}) \cdot \theta^m, \\
\hat{\theta}^t & = \tilde{m} \cdot \hat{\theta}^t + (1 - \tilde{m}) \cdot \theta^t,
\end{align}

To decouple the size of negative samples from the mini-batch size, we employ two negative queues: ${Q}^m = \{k^m_j\}_{j=1}^{N_q}$ and ${Q}^t = \{k^t_j\}_{j=1}^{N_q}$, where each $k_j^m$ and $k_j^t$ is a momentum feature vector extracted via:
\begin{equation}
k_j^m = \hat{f}^m(m_j), \hspace{7pt} k_j^t = \hat{f}^t(t_j).
\end{equation}

Given a training mini-batch $\mathcal{B} = \{(m_i, t_i)\}_{i=1}^{N_b} \subseteq \mathcal{D}$ ($N_b = |\mathcal{B}| \ll N_q$), we compute the momentum features for each sample pair and enqueue them into their respective queues, while simultaneously dequeuing the oldest $N_b$ entries to maintain a fixed queue size. For contrastive learning, each motion sample takes its paired text as the positive example, and all entries in $Q^t$ are treated as negatives. The motion-to-text contrastive loss is formulated as:
\begin{equation}
\label{eqn:m2t-moco}
\mathcal{L}_\text{M2T} \hspace{-1pt}=\hspace{-1pt} - 
\hspace{-3pt} \sum_{i=1}^{N_b} \log \frac{\exp(q^m_i \hspace{-1pt} \cdot k^t_i / \tau)}{\exp(q^m_i \hspace{-1pt} \cdot k^t_i / \tau) \hspace{-2pt}+\hspace{-2pt} \text{neg}(q^m_i, \hspace{-1pt} Q^t, \tau)
},
\end{equation}
where $q^m_i = f^m(m_i)$ and $k^t_i = \hat{f}^t(t_i)$. The negative term is defined by:
\begin{equation}
\text{neg}(q^m_i, Q^t, \tau) = \sum_{k^t_j \in Q^t} \exp(q^m_i \cdot k^t_j / \tau).
\end{equation}
Analogously, we reverse the roles of motion and text to compute the text-to-motion contrastive loss:
\begin{equation}
\label{eqn:t2m-moco}
\mathcal{L}_\text{T2M} \hspace{-1pt}=\hspace{-1pt} -
\hspace{-3pt} \sum_{i=1}^{N_b} \log \frac{\exp(q^t_i \cdot k^m_i / \tau)}{\exp(q^t_i \cdot k^m_i / \tau) \hspace{-2pt}+\hspace{-2pt} \text{neg}(q^t_i, Q^m, \tau)},
\end{equation}
where $q^t_i = f^t(t_i)$ and $k^m_i = \hat{f}^m(m_i)$. The final bidirectional momentum contrastive loss is the sum of both directions:
\begin{equation}
\mathcal{L}_{\text{BMM}} = \mathcal{L}_{\text{M2T}} + \mathcal{L}_{\text{T2M}}.
\end{equation}

\newcommand{\CommentC}[2][gray]{{\color{#1}\textnormal{// #2}}}

\begin{algorithm}[H]
\caption{Bidirectional Momentum Text-Motion Modeling}
\begin{algorithmic}[1]
\Require Training set $\mathcal{D}$
\Require Online encoders $f^t$, $f^m$ with parameters $\theta^t$, $\theta^m$
\Require Momentum encoders $\hat{f}^t$, $\hat{f}^m$ with $\hat{\theta}^t$, $\hat{\theta}^m$
\Require Queues: $Q^t$, $Q^m$
\State Initialize: $\hat{\theta}^t \gets \theta^t$, \quad $\hat{\theta}^m \gets \theta^m$
\While{not converged}
    \State Sample a mini-batch $\mathcal{B} = \{(m_i, t_i)\}_{i=1}^{N_b} \subseteq \mathcal{D}$
    \State \CommentC{Feature encoding}
    \For{each $(m_i, t_i) \in \mathcal{B}$}
        \State $q_i^t \hspace{1.5pt} \gets f^t(t_i)$,\quad $q_i^m \hspace{1.5pt} \gets f^m(m_i)$
        \State $k_i^t \hspace{1.5pt} \gets \hat{f}^t(t_i)$,\quad $k_i^m \gets \hat{f}^m(m_i)$
    \EndFor

    \State \CommentC{Contrastive loss}
    \State Compute $\mathcal{L}_{\text{M2T}}$ and $\mathcal{L}_{\text{T2M}}$ using Eq.(\ref{eqn:m2t-moco}) and Eq.(\ref{eqn:t2m-moco})
    \State $\mathcal{L}_{\text{BMM}} \gets \mathcal{L}_{\text{M2T}} + \mathcal{L}_{\text{T2M}}$

    \State \CommentC{Optimization}
    \State Update $\theta^t, \theta^m$ by minimizing $\mathcal{L}_{\text{BMM}}$

    \State \CommentC{Momentum update}
    \State $\hat{\theta}^t \hspace{4pt} \gets \tilde{m} \cdot \hat{\theta}^t + (1 - \tilde{m}) \cdot \theta^t$
    \State $\hat{\theta}^m \gets \tilde{m} \cdot \hat{\theta}^m + (1 - \tilde{m}) \cdot \theta^m$

    \State \CommentC{Queue update}
    \State Enqueue $\{k_i^t\}_{i=1}^{N_b}$ into $Q^t$, $\{k_i^m\}_{i=1}^{N_b}$ into $Q^m$
    \State Dequeue earliest $N_b$ entries from each queue
\EndWhile
\end{algorithmic}
\end{algorithm}

\subsection{Semantics Spatial-Temporal Attention}      %
\label{sec:spta}

As demonstrated in Figure~\ref{fig:pipeline}(c), for a 2D token map generated by a 2D RVQ-VAE (discussed in later sections), we first add a 2D position encoding~\cite{2dpositionencoding}. The 2D position encoding $P \in \mathbb{R}^{T \times J}$ is obtained by independently applying sinusoidal functions along the temporal and spatial axes. After that, we flatten the 2D token map to a 1D structure, resulting in latent motion vector $z \in \mathbb{R}^{(TJ)\times d}$. Then we extract the text embedding $t \in \mathbb{R}^{1\times d}$ using the text encoder of Clip~\cite{clip}, and get the retrieved text feature $R_t \in \mathbb{R}^{1\times d}$ and retrieved motion feature $R_m \in \mathbb{R}^{1\times d}$ using part-level bmm retriever (as will be explained later). Here, $J$ is the number of joints, $T$ is the number of frames, and $d$ is the feature dimension. Then we perform the adapted attention in the flattened spatial-temporal dimension, as:
\begin{equation}
\label{eqn:spta}
\mathcal{A}_{\text{ssta}}(Q, K, V) = \text{softmax}\left( \frac{{Q} K^T}{\sqrt{d}} + P \right) V,
\end{equation}
\begin{align}
Q &= m , \\
K &= \text{concat}(z, t, [R_m; R_t]), \\
V &= \text{concat}(z, t, R_m).
\end{align}

where $[\cdot;\cdot]$ denotes the concatenation of both terms, $P$ is the 2D Position embedding, and Q, K, V are the query, key, and value matrices.
We refer to this attention mechanism as \textit{Semantics Spatial-Temporal Attention} (SSTA). A simplified variant that only concatenates the text embedding $t$ with the motion vector $z$ and excludes retrieval features, is termed \textit{Spatial-Temporal Attention} (STA). As shown in Figure~\ref{fig:pipeline}(b), both attention mechanisms will be used in the subsequent motion generation module.

\subsection{Training: Part-Level BMM Retriever}
To model fine-grained motion details, we also implement a part-level motion encoder inspired by~\cite{Yu_Tanaka_Fujiwara_2025, parco}. Concretely, we divide the whole-body motion into six parts and embed each part, respectively. These part-level features are then concatenated and reprojected into a latent dimension to produce a fine-grained motion feature, enabling more precise retrieval of part-specific features. For the retriever, we replacing the motion encoder $f^m$ and its corresponding momentum model $\hat{f}^m$ with part-level motion encoder $f^{pl}$ and $\hat{f}^{pl}$ in Equation~\ref{eqn:m2t-moco} and Equation~\ref{eqn:t2m-moco}.
The full training objective of Part-level BMM Retriever is:
\begin{equation}
\label{eqn:pl-bimoco}
\mathcal{L}_{\text{BMM}}^\text{pl} = \mathcal{L}_{\text{M2T}}^\text{pl} + \mathcal{L}_{\text{T2M}}^\text{pl}.
\end{equation}

\subsection{Training: Retrieval-Augmented MoMask}
\subsubsection{Network Architecture}
As illustrated in Figure~\ref{fig:pipeline}(b), ReMoMask includes three key components: a 2D RVQ-VAE encode-decoder quantizes motion into discrete 2D tokens and reconstructs motion from them.  
a 2D retrieval-augmented masked transformer generates base-layer tokens conditioned on text and retrieved features.  
a 2D residual transformer refines the remaining token layers to capture fine-grained details.

\begin{table*}[t]
\centering
\caption{Performance comparison on HumanML3D dataset.}
\label{tab:humanml3d}
\small
\resizebox{\linewidth}{!}{
\begin{tabular}{@{}lccccccc@{}}
\toprule
\textbf{Method} & 
\multicolumn{3}{c}{\textbf{R-Precision $\uparrow$}} & 
\textbf{FID $\downarrow$} & 
\textbf{MM Dist $\downarrow$} & 
\textbf{Diversity $\rightarrow$} & 
\textbf{MultiModality $\uparrow$} \\
\cmidrule(lr){2-4}
& \textit{Top1} & \textit{Top2} & \textit{Top3} & & & & \\ 
\midrule
Real Motions & 0.511 & 0.703 & 0.797 & 0.002 & 2.974 & 9.503 & -- \\
\midrule
MoCoGAN \cite{tulyakov2018mocogan} & 0.037 & 0.072 & 0.106 & 94.41 & 9.643 & 0.462 & 0.019 \\
Dance2Music \cite{lee2019dancing} & 0.033 & 0.065 & 0.097 & 66.98 & 8.116 & 0.725 & 0.043 \\
Language2Pose \cite{ahuja2019language} & 0.246 & 0.387 & 0.486 & 11.02 & 5.296 & 7.676 & -- \\
Text2Gesture \cite{bhattacharya2021text2gestures} & 0.165 & 0.267 & 0.345 & 7.664 & 6.030 & 6.409 & -- \\
T2M \cite{guo2022generating} & 0.457 & 0.639 & 0.740 & 1.067 & 3.340 & 9.188 & 2.090 \\
T2M-GPT \cite{zhang2023t2m} & 0.491 & 0.680 & 0.775 & 0.116 & 3.118 & 9.761 & 1.856 \\
FineMoGen \cite{FineMoGen} &0.504 &0.690 &0.784 &0.151 &2.998 &9.263 &2.696 \\
MDM \cite{tevet2023human} & -- & -- & 0.611 & 0.544 & 5.566 & 9.559 & 2.799 \\
MotionDiffuse \cite{zhang2024motiondiffuse} & 0.491 & 0.681 & 0.782 & 0.630 & 3.113 & 9.410 & 1.553 \\
MoMask~\cite{guo2024momask}  & 0.521 & 0.713 & 0.807 & 0.045 & 2.958 & -- & 1.241 \\
MoGenTS~\cite{yuan2024mogentsmotiongenerationbased} & 0.529 & 0.719 & 0.812 & 0.033 & 2.867 & 9.570 & --   \\
\midrule
ReMoDiffuse \cite{zhang2023remodiffuse} & 0.510 & 0.698 & 0.795 & 0.103 & 2.974 & 9.018 & 1.795 \\
ReMoGPT \cite{yu2024remogpt} & 0.501 & 0.688 & 0.792 & 0.205 & 2.929 & 9.763 & 2.816 \\
RMD \cite{RMD} &0.524 &0.715 &0.811 &0.111 &2.879 & 9.527 &2.604 \\
MoRAG-Diffuse \cite{morag} &0.511 &0.699 &0.792 &0.270 &2.950 &9.536 &2.773 \\
\midrule
\textbf{ReMoMask (Ours)} & \textbf{0.531} & \textbf{0.722} & \textbf{0.813} & 0.099 & \textbf{2.865} & 9.535 & \textbf{2.823} 
\cr
\bottomrule
\end{tabular}
}
\end{table*}

\subsubsection{2D Residual VQ-VAE}
Given a motion $\mathbf{m} \in \mathbb{R}^{T \times J}$, we first extract 2D latent features $\hat{\mathbf{y}} \in \mathbb{R}^{T \times J}$ using a 2D convolutional encoder $E_{2d}$. We then apply residual vector quantization (RVQ)~\cite{guo2024momask} with $L{+}1$ levels:
\begin{equation}
\mathbf{y}^l = Q^l(\mathbf{r}^l), \quad \mathbf{r}^{l+1} = \mathbf{r}^l - \mathbf{y}^l,
\end{equation}
starting from $\mathbf{r}^0 = \hat{\mathbf{y}}$, where $Q^l(\cdot)$ denotes the vector quantization operation at level $l$, mapping each latent vector to its nearest code in a learnable codebook. The final summed quantized representation $\sum_{l=0}^L \mathbf{y}^l$ is then fed into a 2D convolutional decoder $D_{2d}$ to reconstruct motion, resulting in $\mathbf{\hat{m}} \in \mathbb{R}^{T \times J}$.

We train the model by minimizing reconstruction and embedding losses using the straight-through gradient estimator, as
\begin{equation}
\mathcal{L}_{\text{2D-RVQ}} = \|\mathbf{m} - \hat{\mathbf{m}}\|_1 + \gamma\sum_{l=1}^{L} \|\mathbf{r}^l - \text{sg}[\mathbf{y}^l]\|_2^2,
\end{equation}
where $\text{sg}[\cdot]$ denotes the stop-gradient operation, and $\gamma$ controls the strength of the embedding loss.

\subsubsection{2D Retrieval-Augmented Masked-Transforemr}
Under the hypothesis of hierarchical RVQ-VAE, the base quantization layer—generated by a masked transformer—captures the coarse motion structure, while the residual transformer layers refine fine-grained details. Since the base layer captures the main semantics of motion, we introduce retrieval-augmented context to this stage only, aiming to enhance structural reconstruction without overburdening the refinement stages.

We design our 2D retrieval-augmented masked transformer to generate the base-layer motion tokens $\mathbf{y}^0 \in \mathbb{R}^{T \times J}$, conditioned on the text embedding $t$, retrieved text feature $R_t$, and retrieved motion feature $R_m$, all of which are fused using SSTA. To train the model, we randomly mask a subset of tokens in $\mathbf{y}^0$, replacing them with a $[\mathrm{MASK}]$ token to obtain a corrupted sequence  $\mathbf{y}^0_{\text{msk}}$. We perform a 2D Mask strategy~\cite{yuan2024mogentsmotiongenerationbased} by first randomly masking along the temporal dimension and then randomly masking along the spatial dimension on those unmasked frames. The model is then trained to reconstruct the original tokens by minimizing the negative log-likelihood:
\begin{align}
\mathcal{L}_{\text{mask}}^{\text{rag}} = \sum_{\text{[MASK]}} -\log p(\mathbf{y}^0 \mid \mathbf{y}_{msk}^0, t, R_t, R_m).
\end{align}
We also employ a masking ratio schedule and a BERT-style remasking strategy, extending the previous method~\cite{maskgit, muse, bert, guo2024momask}. 

\subsubsection{2D Residual Transformer}
The architecture of our 2D residual transformer mirrors that of the 2D retrieval augmented masked transformer, except that we adopt STA instead of SSTA. During training, we randomly select a quantization layer $l \in [1, L]$ to predict. All tokens from previous layers $\mathbf{y}^{0:l{-}1}$ are summed to form the latent input, together with the quantizer layer index $l$ and the text condition $t$. The model is optimized by minimizing the negative log-likelihood:
\begin{equation}
\mathcal{L}_{\text{res}} = \sum_{l=1}^{L} -\log p\left(\mathbf{y}^{l} \mid \mathbf{y}^{0:l{-}1}, l, t \right),
\end{equation}

\begin{table*}[t]
\centering
\caption{Performance comparison on KIT-ML dataset.}
\label{tab:kitml}
\small
\resizebox{\linewidth}{!}{
\begin{tabular}{@{}lccccccc@{}}
\toprule
\textbf{Method} & 
\multicolumn{3}{c}{\textbf{R-Precision $\uparrow$}} & 
\textbf{FID $\downarrow$} & 
\textbf{MM Dist $\downarrow$} & 
\textbf{Diversity $\rightarrow$} & 
\textbf{MultiModality $\uparrow$} \\
\cmidrule(lr){2-4}
& \textit{Top1} & \textit{Top2} & \textit{Top3} & & & & \\ 
\midrule
Real Motions & 0.424 & 0.649 & 0.779 & 0.031 & 2.788 & 11.08 & -- \\
\midrule
MoCoGAN \cite{tulyakov2018mocogan} & 0.022 & 0.042 & 0.063 & 82.69 & 10.47 & 3.091 & 0.250 \\
Language2Pose \cite{ahuja2019language} & 0.221 & 0.373 & 0.483 & 6.545 & 5.147 & 9.073 & -- \\
Dance2Music \cite{lee2019dancing} & 0.031 & 0.058 & 0.086 & 115.4 & 10.40 & 0.241 & 0.062 \\
Text2Gesture \cite{bhattacharya2021text2gestures} & 0.156 & 0.255 & 0.338 & 12.12 & 6.964 & 9.334 & -- \\
T2M \cite{guo2022generating} & 0.370 & 0.569 & 0.693 & 2.770 & 3.401 & 10.91 & 1.482 \\
MotionDiffuse \cite{zhang2024motiondiffuse} & 0.417 & 0.621 & 0.739 & 1.954 & 2.958 & 11.10 & 0.730 \\
T2M-GPT \cite{zhang2023t2m} & 0.416 & 0.627 & 0.745 & 0.514 & 3.007 & 10.92 & 1.570 \\
MDM \cite{tevet2023human} & -- & -- & 0.396 & 0.497 & 9.191 & 10.85 & 1.907 \\
MoMask~\cite{guo2024momask} & 0.433 & 0.656 & 0.781 & 0.204 &2.779 & -- & 1.131 \\  
MoGenTS~\cite{yuan2024mogentsmotiongenerationbased} & 0.445 & 0.671 & 0.797 & 0.143 & 2.711 & 10.918 & --   \\
\midrule
ReMoDiffuse \cite{zhang2023remodiffuse} & 0.427 & 0.641 & 0.765 & 0.155 & 2.814 & 10.80 & 1.239 \\
\midrule
\textbf{ReMoMask (Ours)} & \textbf{0.453} & \textbf{0.682} & \textbf{0.805} & \textbf{0.138} & \textbf{2.682} & 10.83 & \textbf{2.017} 
\cr
\bottomrule
\end{tabular}
}
\end{table*}

\subsection{Inference}
In the inference, we start from an empty 2D token map that all tokens are masked, defined as $\mathbf{y} \in  \mathbb{R}^{T \times J}$. Then the 2D retrieval-augmented masked transformer repeatedly predicts the masked tokens as a base quantization layer by N iterations, conditioned on the caption embedding $t$, the retrieved motion feature $R_m$,  and the retrieved text features $R_t$. Once the prediction is completed, the 2D residual transformer progressively predicts the residual tokens of the rest quantization layers. As a final stage, all tokens are decoded and projected back to motion sequences through the 2D RVQ-VAE decoder.

\subsubsection{RAG Classifier Free Guidance}
We extend the classifier-free guidance (CFG) to incorporate the text embedding $t$, retrieved text feature $R_t$, and retrieved motion feature $R_m$ as conditional inputs. Explicitly, we define the guidance condition as $\{t, R_t, R_m\}$, the final logits are computed as:
\begin{equation}
    \text{logits} = (1+s) \cdot \text{logits}_\text{con} - s \cdot \text{logits}_\text{un},
\end{equation}
\begin{equation}
    \text{con} = \{t, R_t, R_m\} 
\end{equation}
where logits is the output of the final linear projection layer and $s$ is the guidance scale (set to 4).

During training, unconditional sampling is applied with a probability of $10\%$ to enable guidance-free learning. At inference, R-CFG is applied to the final projection layer prior to softmax.

\section{Experiment}

\subsection{Dataset and Evaluation Metrics}
We evaluate our model on HumanML3D~\cite{HumanML3D} and KIT-ML~\cite{KIT-ML} datasets. The HumanML3D dataset~\cite{HumanML3D} stands as the largest available dataset focused solely on 3D body motion and associated textual descriptions. It consists of 14616 motion sequences and 44970 text descriptions, and KIT-ML consists of 3911 motions and 6278 texts. The motion pose is extracted into the motion feature with dimensions of 263 and 251 for HumanML3D and KIT-ML respectively. Following previous methods~\cite{HumanML3D}, the datasets are augmented by mirroring,and divided into training, testing,and validation sets with the ratio of 0.8:0.15:0.05.

\subsubsection{Evaluation Metrics} We adapt standard evaluation metrics to assess various aspects of our experiments:  For overall motion quality, we propose Fréchet Inception Distance (FID) to measure the distributional difference between high-level features of generated and real motions. For semantic alignment between input text and generated motions, we propose R-Precision and multimodal distance. For diversity of motions generated from the same text, we propose Multimodality.

\subsection{Implementation Details}

Our framework is trained on four NVIDIA H20 GPUs using PyTorch with a batch size of 256 and a learning rate of $2\times10^{-4}$. For motion quantization, we employ a \textbf{2D RVQ-VAE} structure following~\cite{yuan2024mogentsmotiongenerationbased}. The pose data is restructured into a joint-based format of size $12\times J$ and quantized into 2D latent representations. This is achieved using two codebooks: (1) a joint VQ codebook containing 256 codes (dimension 1024), and (2) a global VQ codebook with 256 codes (dimension 1024) to capture holistic motion information.For motion generation, we implement two transformer architectures:M-trans (Motion Transformer): 6 layers, 8 attention heads, and 512 latent dimensions, R-trans (Residual Transformer): 6 layers, 6 attention heads, and 384 latent dimensions. The \textbf{2D Residual Transformer} structure follows~\cite{guo2024momask} with 5 residual layers.For retrieval enhancement, we design a \textbf{2D Retrieval-augmented Masked Transformer} using momentum contrastive learning. The motion encoder adopts RemoGPT's 4-layer Transformer architecture~\cite{yu2024remogpt} with 512 latent dimensions. Text embeddings are generated using DistilBERT~\cite{distillbert}. Key hyperparameters include: momentum coefficient $m=0.99$, temperature $\tau=0.07$, and dynamic queue size 65536. This module is trained on eight NVIDIA A800 GPUs with batch size 128 for 200 epochs.

\subsection{Main Results}
\subsubsection{Evaluation of Motion Generation}
we compare our model with previous text-to-motion works, including combing RAG method and without RAG method. From the results reported in Table~\ref{tab:humanml3d},~\ref{tab:kitml}, our method outperforms all previous methods on both the HumanML3D and the KIT-ML datasets, which demonstrates the effectiveness of our method. Crucially, the FID is decreased by 0.093 on HumanML3D compared and is decreased by 0.066 on KIT-ML compared to MoMask~\cite{guo2024momask}. Moreover, the R-precision even significantly surpasses the ground truth.

\subsubsection{Evaluation of Retriever}  
As shown in Table~\ref{tab:uni/bi_moco}, in the \textit{text-to-motion retrieval} task, BMM achieves state-of-the-art performance with significant improvements across key metrics. It obtains the highest scores in R1 (13.76), R2 (21.03), R3 (25.63), and R5 (32.40), outperforming PL-TMR with absolute gains ranging from 2.76\% to 2.92\%. Although its R10 score (43.27) is slightly lower than that of PL-TMR (43.43), the overall superiority of BMM is evident. Similarly, BMM demonstrates strong performance in the \textit{motion-to-text retrieval} task, achieving the highest results in R1 (14.80) and R3 (25.60), with a notable 2.55\% absolute improvement in R1 over PL-TMR. However, it shows limitations in higher-recall metrics: its R5 (25.75) and R10 (34.61) lag behind PL-TMR by 2.59\% and 4.50\%, respectively. Moreover, BMM's MedR (25.00) also underperforms, suggesting reduced effectiveness in retrieving multiple relevant texts per motion.

\subsection{Ablation Study}    
This section conducts systematic ablation experiments to validate the contributions of ReMoMask's core modules: Bidirectional Momentum Model (BMM), Semantic Spatio-Temporal Attention (SSTA), Retrieval-Augmented Classifier-Free Guidance (RAG-CFG), and local retrieval mechanism. Results conclusively demonstrate the necessity of each innovative component.

\subsubsection{Core Module Contribution Analysis}
As shown in Table~\ref{tab:ablation1}, the full model (ReMoMask) achieves comprehensive SOTA performance on HumanML3D: 
\textit{BMM Module}: Removal causes 16.2\% Top1 R-Precision drop (0.531→0.445) and 50.18\% FID degradation (0.411→0.825), proving its irreplaceability in cross-modal alignment. \textit{SSTA Module}: Replacement with feature concatenation leads to 61.2\% multimodality collapse (2.823→1.094) and 6.1\% MM Dist increase (2.865→3.04), highlighting its critical role in motion diversity. \textit{RAG-CFG}: Deactivation reduces Top1 R-Precision by 22.6\% (0.531→0.411), confirming its efficacy in enhancing text-motion consistency. \textit{Local Retrieval}: Global retrieval substitution decreases Top3 R-Precision by 9.8\% (0.813→0.733) and diversity by 4.8\% (9.535→9.08), demonstrating the superiority of local context retrieval.

\subsubsection{Bidirectional Momentum Queue Optimization}
Table~\ref{tab:ablation2} reveals the impact of momentum queue design on cross-modal retrieval: \textit{Text→Motion Retrieval}: Bidirectional queues improve R1 by 31.3\% (10.48→13.76) and reduce MedR by 15.8\% (19→16) compared to no queues. \textit{Motion→Text Retrieval}: Unidirectional queues cause catastrophic failure (R1=0.70), while bidirectional queues boost R1 by 41.0\% (10.50→14.80), proving symmetric negative sample queues are indispensable for bidirectional retrieval.
The optimal configuration significantly outperforms baselines in both tasks, validating the robustness of our bidirectional momentum design.

\begin{table}[t]
\centering
\caption{Performance comparison of text-motion retrieval tasks on HumanML3D dataset.}
\label{tab:uni/bi_moco}
\small
\resizebox{\columnwidth}{!}{
\begin{tabular}{@{}lcccccccc@{}}
\toprule
\textbf{Method} & \textbf{Params} & \textbf{R1↑} & \textbf{R2↑} & \textbf{R3↑} & \textbf{R5↑} & \textbf{R10↑} & \textbf{MedR↓} \\
\midrule
\multicolumn{8}{c}{\textbf{Text-to-motion retrieval}} \\
\midrule
TMR & 82M & 8.92 & 12.04 & 16.33 & 22.06 & 33.37 & 25.00 \\
MotionPatches & 152M & 10.80 & 14.98 & 20.00 & 26.72 & 38.02 & 19.00 \\
PL-TMR & 118M & 11.00 & 17.02 & 22.18 & 29.48 & 43.43 & 14.00 \\
\midrule
\textbf{BMM} & 238M & \textbf{13.76} & \textbf{21.03} & \textbf{25.63} & \textbf{32.40} & 43.27 & 16.00 \\
\midrule
\multicolumn{8}{c}{\textbf{Motion-to-text retrieval}} \\
\midrule
TMR & 82M & 9.44 & 11.84 & 16.90 & 22.92 & 32.21 & 26.00 \\
MotionPatches & 152M & 11.25 & 13.86 & 19.98 & 26.86 & 37.40 & 20.00 \\
PL-TMR & 118M & 12.25 & 14.95 & 21.45 & 28.34 & 39.11 & 19.00 \\
\midrule
\textbf{BMM} & 238M & \textbf{14.80} & \textbf{15.63} & \textbf{25.60} & 25.75 & 34.61 & 25.00 \\
\bottomrule
\end{tabular}
}
\end{table}

\begin{table}[!ht]
\centering
\caption{\label{tab:ablation1}
Ablation study 1 on HumanML3D dataset.
We test ReMoMask's core modules: BMM, SSTA, RAG-CFG, and local retrieval mechanism.
}
\small
\resizebox{\linewidth}{!}{
\begin{tabular}{@{}lccccccc@{}}
\toprule
\textbf{Method} & 
\multicolumn{3}{c}{\textbf{R-Precision $\uparrow$}} & 
\textbf{FID $\downarrow$} & 
\textbf{MM Dist $\downarrow$} & 
\textbf{Diversity $\rightarrow$} & 
\textbf{MultiModality $\uparrow$} \\
\cmidrule(lr){2-4}
& \textit{Top1} & \textit{Top2} & \textit{Top3} & & & & \\ 

\midrule
w/o BMM  & 0.445 & 0.639 & 0.751 & 0.825 & 3.44 & 8.80 & 1.017 \\
w/o SSTA  & 0.495 & 0.652 & 0.789 & 0.714 & 3.04 & 9.39 & 1.094 \\
w/o RAG-CFG  & 0.411 & 0.612 & 0.741 & 0.798 & 3.16 & 9.12 & 1.088 \\
w/o Coarse-Level Retrieval  & 0.402 & 0.644 & 0.733 & 0.722 & 3.32 & 9.08 & 1.044\\
\midrule
\textbf{ReMoMask (Ours)} & \textbf{0.531} & \textbf{0.722} & \textbf{0.813} & 0.411 & \textbf{2.865} & 9.535 & \textbf{2.823} \\
\bottomrule
\end{tabular}
}
\end{table}
\begin{table}[!ht]
\centering
\caption{
\label{tab:ablation2}
Ablation study 2 on HumanML3D.We explored the importance of BMM}
\small
\resizebox{\columnwidth}{!}{
\begin{tabular}{cc|cccccccc}
\toprule
\textbf{text queue} & \textbf{motion queue} & \textbf{R1↑} & \textbf{R2↑} & \textbf{R3↑} & \textbf{R5↑} & \textbf{R10↑} & \textbf{MedR↓} \\
\midrule
\multicolumn{8}{c}{\textbf{Text-to-motion retrieval}} \\
\midrule
\xmark & \xmark & 10.48 & 15.80 & 20.90 & 28.59 & 41.25 & 19.00 \\
\xmark & \cmark & 12.44 & 18.98 & 22.79 & 29.72 & 43.02 & 17.00 \\
\cmark & \cmark & 13.76 & 21.03 & 25.63 & 32.40 & 43.27 & 16.00 \\
\midrule
\multicolumn{8}{c}{\textbf{Motion-to-text retrieval}} \\
\midrule
\xmark & \xmark & 10.50 & 13.36 & 18.64 & 24.98 & 36.83 & 20.00 \\
\xmark & \cmark & 0.70 & 1.02 & 1.22 & 1.87 & 3.18 & 552.00 \\
\cmark & \cmark & 14.80 & 15.63 & 25.60 & 25.75 & 34.61 & 25.00 \\
\bottomrule
\end{tabular}
}
\end{table}

\section{Conclusion}
In this paper, we introduce an innovative retrieval-augmented masked model, ReMoMask, for text-driven motion generation. The proposed bidirectional momentum text-motion relation modeling enlarges the set of negative samples across modalities, facilitating more effective contrastive learning for the part-level retriever. Quantizing the motion sequence into a 2D token map and applying well-designed cross-attention with textual and retrieved conditions enables a more expressive fusion of conditional semantics and spatio-temporal motion dynamics. Extensive experiments on HumanML3D and KIT datasets demonstrate that our model achieves SOTA performance.

\clearpage
\appendix

\section{User Study}
To comprehensively evaluate the generation capability of \textbf{ReMoMask}, we conducted a comparative user study.  
We randomly selected 20 text prompts from the HumanML3D test set and generated motion sequences using \textbf{ReMoMask}, current state-of-the-art retrieval-augmented method (ReMoDiffuse), generative model (MoMask), and ground truth motions.

\begin{figure*}
        \centering
    \includegraphics[width=1\linewidth]{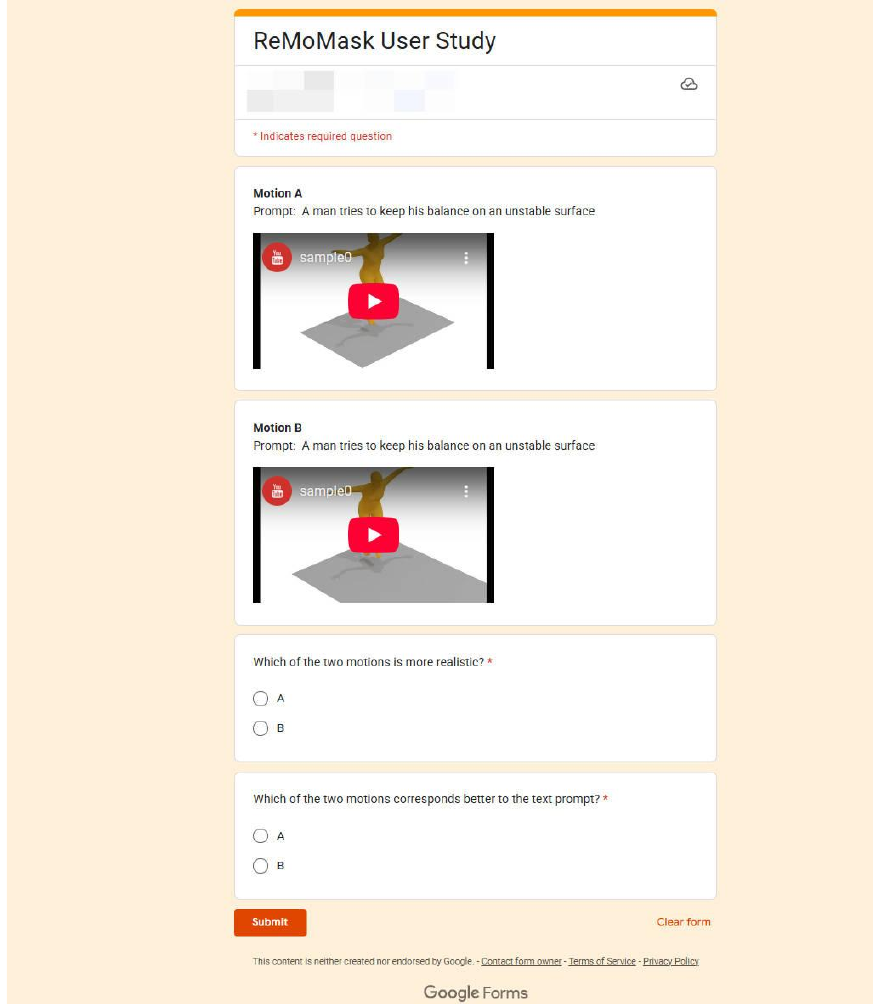}
    \caption{
    This figure illustrates the User Interface (UI) used in the ReMoMask User Study. Participants are presented with two motion videos, labeled as Motion A and Motion B, alongside a shared textual prompt. The motion clips are sampled from outputs generated by different models or the ground truth (GT), with model identities anonymized and video order randomized. Participants are asked to answer two evaluative questions: (1) “Which of the two motions is more realistic?”, assessing the visual plausibility and motion quality; and (2) “Which of the two motions corresponds better to the text prompt?”, evaluating the semantic alignment between the motion and the given description. This dual-question design enables a comprehensive human assessment of both motion realism and text-motion correspondence.
    }
    \label{fig:UI}
\end{figure*}

We employ a forced-choice paradigm in our user study, asking participants two key questions: “Which of the two motions is more realistic?” and “Which of the two motions corresponds better to the text prompt?”. The study is conducted via a Google Forms interface, as illustrated in Figure~\ref{fig:UI}. To ensure fairness and reduce potential bias, the names of the generative models are hidden, and the order of presentation is randomized for each question. In total, over 50 participants took part in the evaluation.

\begin{figure}[h]
    \centering
    \includegraphics[width=\linewidth]{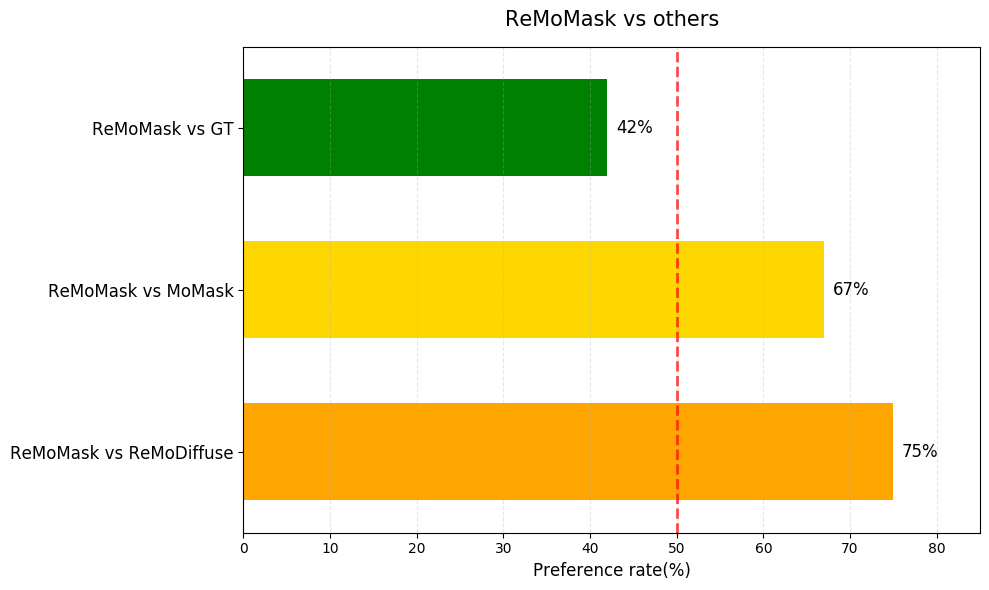}
    \caption{
      Motion Quality User Study
    }
    \label{fig:picture1}
\end{figure}

\begin{figure}[h]
    \centering
    \includegraphics[width=\linewidth]{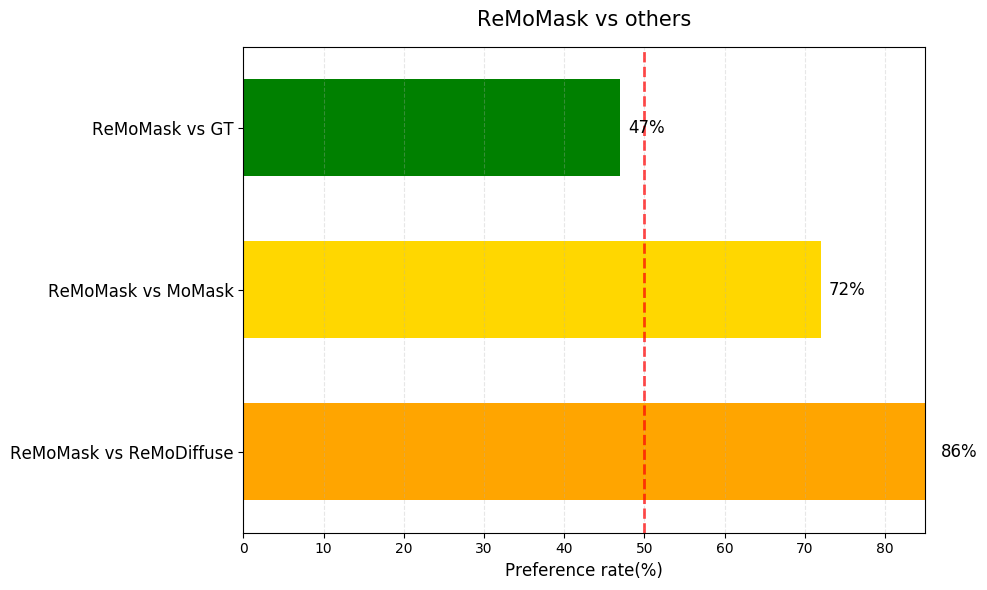}
    \caption{
    Text-Motion Correspondence User Study
    }
    \label{fig:picture2}
\end{figure}

Empirical results, depicted in Figure~\ref{fig:picture1} and Figure~\ref{fig:picture2}, underscore ReMoMask’s strong capability to generate motions that are not only realistic but also closely aligned with textual descriptions. Specifically, as shown in Figure~\ref{fig:picture1}, ReMoMask achieves a 42\% preference rate over ground truth (GT) in terms of realism. Although GT motions are derived from real human data, this result indicates that ReMoMask is perceived as comparably realistic by human evaluators. Moreover, the model significantly outperforms both baselines: it achieves 67\% preference over MoMask and 75\% over ReMoDiffuse, demonstrating its strength in producing high-quality, lifelike motion sequences.

In terms of text correspondence (reported in Figure~\ref{fig:picture2}), ReMoMask attains a 47\% preference rate over GT, suggesting that its generated motions exhibit nearly human-level alignment with text prompts. Compared to the baselines, ReMoMask again shows substantial improvements, with 72\% preference over MoMask and 86\% over ReMoDiffuse.

\section{Limitation and Future Work}
\textbf{Current Limitations}: BMM's dual queues and SSTA's 2D attention significantly increase the model parameters (238M), hindering real-time deployment. Furthermore, experiments conducted primarily on short sequences (<100 frames) lack validation for complex motions requiring strong spatiotemporal coherence, such as dance. Part-level retrieval also struggles with abstract textual descriptions (e.g., "jumping joyfully") due to reliance on predefined motion partitions. Additionally, generated motions may violate biomechanical constraints (e.g., joint rotation limits) due to the lack of physics-based verification.

\textbf{Proposed Future Work}: To address these limitations, we propose: (1) Adopting knowledge distillation or sparse attention mechanisms to reduce model size; (2) Decomposing long motions into sub-actions and applying phased SSTA to enhance temporal consistency; (3) Integrating Large Language Models (LLMs, e.g., GPT-4) to parse abstract texts and dynamically adapt part-level retrieval; (4) Incorporating physical constraint losses during RVQ-VAE decoding to ensure biomechanically valid motions.

\section{Visualization}
\begin{figure*}
        \centering
    \includegraphics[width=1\linewidth]{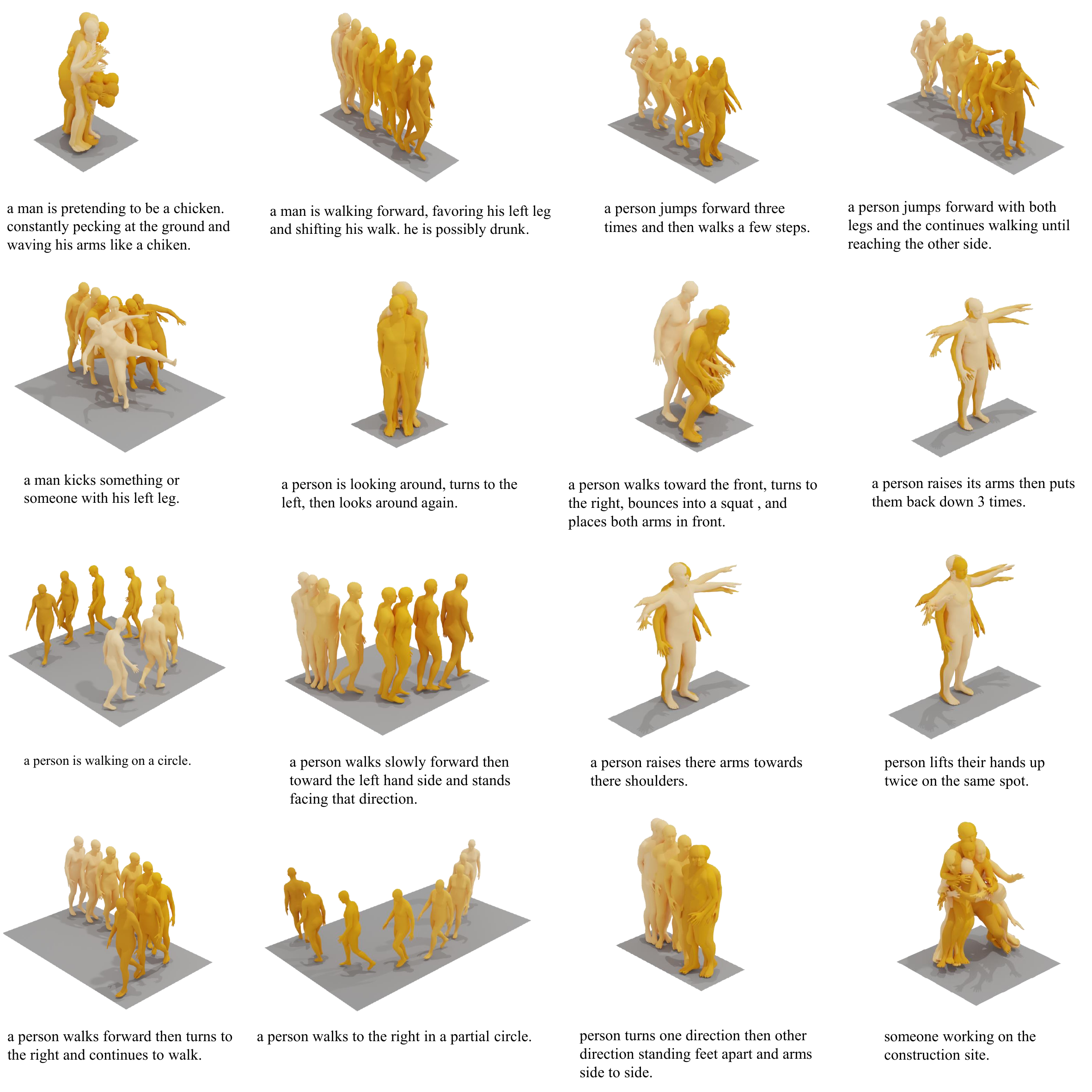}
    \caption{
We randomly sample and visualize 16 motions generated by the proposed ReMoMask framework. These examples are conditioned on diverse prompts randomly selected from the HumanML3D~\cite{HumanML3D}, providing qualitative evidence of the model’s ability to synthesize a wide range of realistic and semantically coherent motions.
    }
    \label{fig:demo}
\end{figure*}

Figure~\ref{fig:demo} demonstrates our model's capability in generating diverse human motions. The 16 randomly inferred samples exhibit complex motion patterns such as directional transitions ("walks toward the front, turns to the right"), rhythmic actions ("raises arms three times"), and semantically rich behaviors ("pretending to be a chicken"). This showcases our model's proficiency in capturing nuanced motion dynamics and temporal transitions.

\begin{figure*}
        \centering
    \includegraphics[width=1\linewidth]{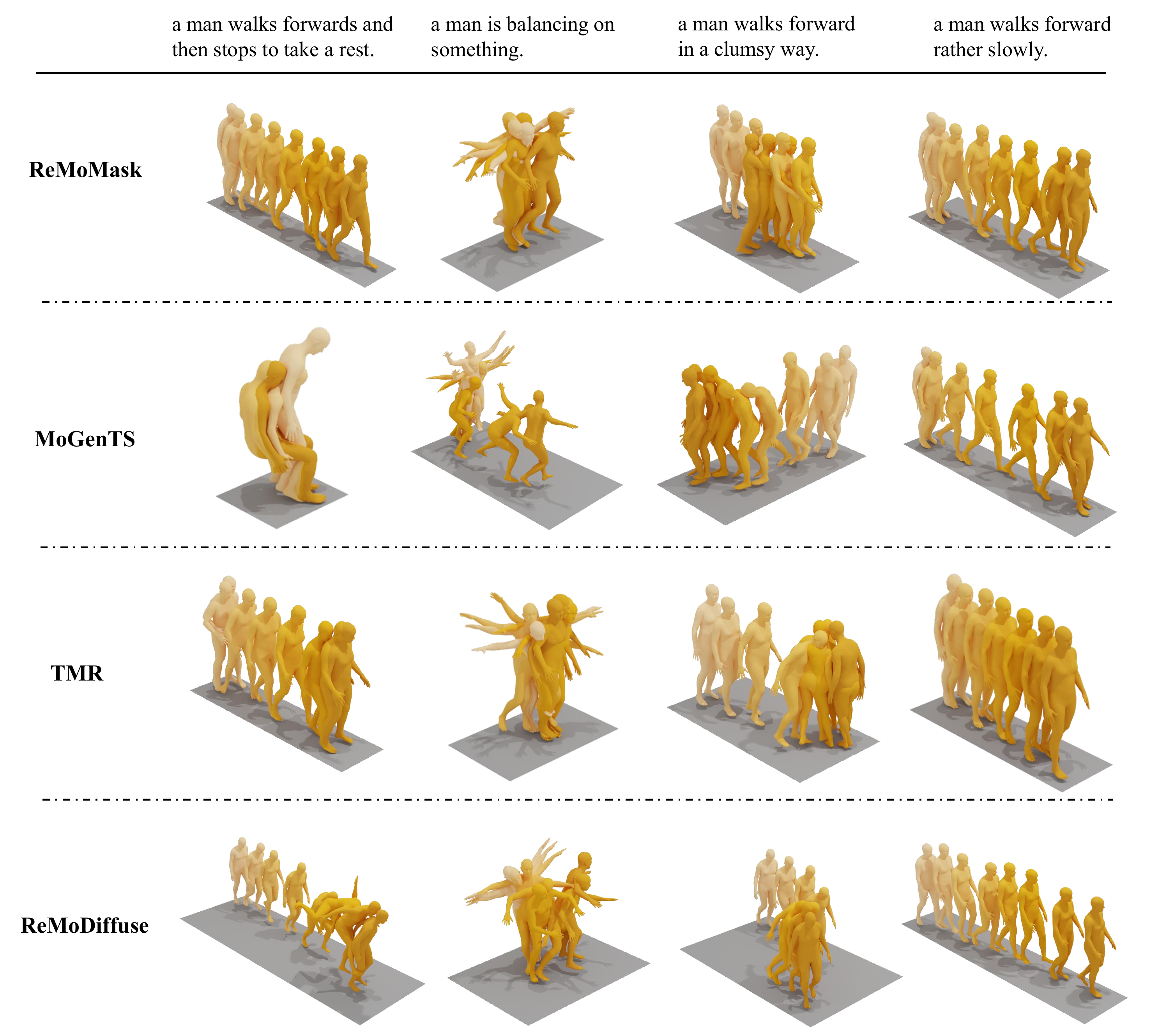}
    \caption{
    Comparison of the proposed ReMoMask with three state-of-the-art methods: MoGenTS~\cite{yuan2024mogentsmotiongenerationbased}, TMR~\cite{petrovich2023tmrtexttomotionretrievalusing}, and ReMoDiffuse~\cite{zhang2023remodiffuse} We visualize motion sequences generated in response to three distinct text prompts. Each row corresponds to a specific prompt, and each column represents the output of a different method. The results demonstrate that ReMoMask produces more realistic and semantically aligned motions compared to existing approaches.
    }
    \label{fig:demo1}
\end{figure*}

Figure~\ref{fig:demo1} provides a comparative analysis against MeGenTS, TMR, and ReMoDiffuse. While baseline models generate basic motions like walking or balancing, our approach consistently produces more natural transitions (e.g., "walks forward then turns" vs. simple linear motion) and physically plausible sequences (e.g., multi-step "jumps forward three times"). The visual comparison highlights our model's superior handling of motion complexity and behavioral expressiveness.

\end{document}